\documentclass[letterpaper]{article} 
\usepackage{aaai25}  
\usepackage{times}  
\usepackage{helvet}  
\usepackage{courier}  
\usepackage[hyphens]{url}  
\usepackage{graphicx} 
\usepackage{natbib}  
\usepackage{caption} 
\usepackage{algorithm}
\usepackage{algorithmic}
\usepackage{amsmath}
\usepackage{amssymb}
\usepackage{booktabs} 
\usepackage{arydshln}
\usepackage{newfloat}
\usepackage{multirow}
\usepackage{color}
\usepackage{makecell}
\usepackage{listings}
\DeclareCaptionStyle{ruled}{labelfont=normalfont,labelsep=colon,strut=off} 
\floatstyle{ruled}
\newfloat{listing}{tb}{lst}{}
\floatname{listing}{Listing}
\makeatletter
\renewcommand{\@fnsymbol}[1]{%
  \ifcase#1\or †\or *\or §\or ‖\or ‡\or **\else\@ctrerr\fi}
\makeatother
\title{SGTC: Semantic-Guided Triplet Co-training for Sparsely Annotated Semi-Supervised Medical Image Segmentation}
\author{
    Ke Yan\textsuperscript{\rm 1}\equalcontrib,
    Qing Cai\textsuperscript{\rm 1}\equalcontrib,
    Fan Zhang\textsuperscript{\rm 2}\footnote{Corresponding authors.},
    Ziyan Cao\textsuperscript{\rm 1},
    Zhi Liu\textsuperscript{\rm 3}\footnotemark[2]\\
}
\affiliations{
    \textsuperscript{\rm 1}Faculty of Computer Science and Technology, Ocean University of China\\
    \textsuperscript{\rm 2}School of Automation, Northwestern Polytechnical University\\
    \textsuperscript{\rm 3}School of Information Science and Engineering, Shandong University\\
    \{yk6923, caoziyan\}@stu.ouc.edu.cn, cq@ouc.edu.cn, zfnlxx@mail.nwpu.edu.cn, liuzhi@sdu.edu.cn

}

\usepackage{bibentry}

\begin{document}

\maketitle

\begin{abstract}
  Although semi-supervised learning has made significant advances in the field of medical image segmentation, fully annotating a volumetric sample slice by slice remains a costly and time-consuming task. Even worse, most of the existing approaches pay much attention to image-level information and ignore semantic features, resulting in the inability to perceive weak boundaries. To address these issues, we propose a novel \underline{S}emantic-\underline{G}uided \underline{T}riplet \underline{C}o-training (SGTC) framework, which achieves high-end medical image segmentation by only annotating three orthogonal slices of a few volumetric samples, significantly alleviating the burden of radiologists. Our method consist of two main components. Specifically, to enable semantic-aware, fine-granular segmentation and enhance the quality of pseudo-labels, a novel semantic-guided auxiliary learning mechanism is proposed based on the pretrained CLIP. In addition, focusing on a more challenging but clinically realistic scenario, a new triple-view disparity training strategy is proposed, which uses sparse annotations (i.e., only three labeled slices of a few volumes)  to perform co-training between three sub-networks, significantly improving the robustness. Extensive experiments on three public medical datasets demonstrate that our method outperforms most state-of-the-art semi-supervised counterparts under sparse annotation settings. The source code is available at https://github.com/xmeimeimei/SGTC.
      
\end{abstract}

%

\section{Introduction}

\begin{figure}[!t]
\footnotesize
\centering
\setlength{\abovecaptionskip}{0cm}
  \includegraphics[width=1\linewidth]{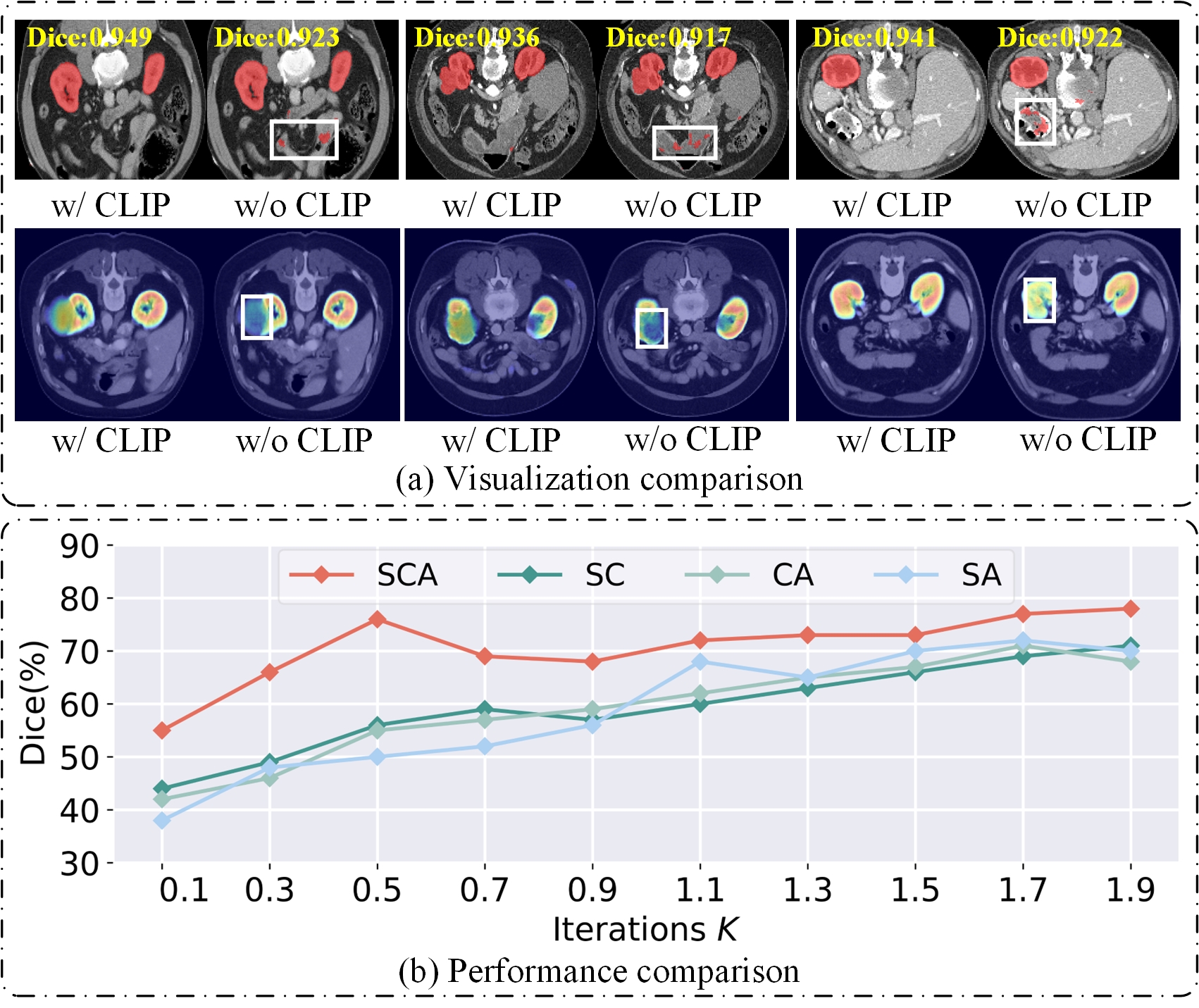}
  \caption{(a) The semantic guidance in visualization comparison. (b) The performance under different sparse annotation strategy. The S, C, and A indicate sagittal, coronal, and axial annotated slices.}
  \label{fig_1}
\end{figure}

Segmentation of anatomical structures and pathology within medical images holds paramount importance for clinical diagnosis \cite{zhou2019semi}, treatment planning~\cite{li2023scribblevc,zhang2024federated}, and disease research~\cite{zhang2022pixelseg}. While significant progress has been achieved through deep learning-based segmentation techniques, many approaches encounter substantial bottlenecks when lacking sufficient well-annotated datasets \cite{zhang2024cross,zhang2022rvlsm}. Consequently, there is a critical need to develop more effective yet precise segmentation methods to decrease the dependence on large-scale pixel-wise annotated data.

Considering that unlabeled data are easy to obtain, semi-supervised medical image segmentation has emerged as the predominant strategy in medical domain, utilizing a limited number of labeled data and lots of unlabeled data~\cite{wang2024new,wang2024towards,zhang2022probabilistic,cai2018aadaptive,zhou2023unified}. 
Current semi-supervised medical segmentation methods can be classified into two categories. The first one is pseudo-label-based methods \cite{bai2017semi,uamt,bai2023bidirectional}, which estimate the pseudo labels based on a few labeled samples and then attach them to unlabeled ones followed by a fine-tuning stage on the newly enlisted training set. The other mainstream approaches are consistency regularization \cite{chen2023magicnet,ucmt,li2020transformation}, which enforce the output consistency for the inputs under different image or feature perturbations.

Despite the progress, these methods need dense annotations, requiring to fully annotate the entire volume of limited labeled data slice by slice, it remains a costly and time-consuming task due to the considerable number of slices per volume. Generally, radiologists only annotate a few slices from the volumetric medical data and leave others unlabeled~\cite{zhang2024dslsm,cai2019saliency,cai2021novel,yun2023uni,gcl}.
Under this condition, the semi-supervised methods based on sparse annotations have been proposed, for example, PLN~\cite{li2022pln} annotates only one slice per volume and propagates the pseudo labels for other slices using the parasitic-like network. Similarly, Desco~\cite{Desco} annotates two orthogonal slices per volume and utilizes the SyNRA method from the Ants~\cite{avants2011reproducible} library for registration, constructing consistency between pseudo labels generated from two perspectives. Nonetheless, these methods have limitations in semantic understanding, which hinders the network's ability to accurately recognize anatomical structures and lesions. As shown in Figure \ref{fig_1}(a), without semantic information providing contextual clues, important features can be lost, resulting in imprecise segmentation results. Furthermore, slice-by-slice propagation or registration is a very time-consuming process, especially when the volume comprises a substantial number of slices. Worse still, annotating only one or two slices misses complementary information from three different views, resulting in an incomplete representation of the volume's data distribution, consequently leading to a suboptimal segmentation (See Figure \ref{fig_1}(b)). 

To overcome these challenges, this paper proposes a novel semantic-guided triplet co-training framework, dubbed SGTC, which achieves semantic-aware and fine-granular semi-supervised medical image segmentation by merely annotating three orthogonal slices of a few volumetric samples. Specifically, to enable the use of text representations to connect semantic-aware features, a novel semantic-guided auxiliary learning mechanism is proposed. It enhances pseudo-label quality for abundant unlabeled medical data by refining the intricate structures and weak boundaries. Besides, to better align with the spatial information distribution of volumetric data using sparse annotations, a novel triple-view disparity training strategy is proposed, which better maintains the disparity of sub-networks during training, allowing the sub-networks to learn complementary knowledge from each other. More importantly, it focuses on a more challenging but clinically realistic scenario, where radiologists just need to annotate three orthogonal slices of a few volumetric samples. Extensive experiments on LA2018, KiTS19, and LiTS datasets under sparse annotation settings show that our SGTC achieves superior performance against most state-of-the-art semi-supervised learning methods.

The primary contributions of this paper include: 
\begin{itemize}
	\item 
	A novel semantic-guided auxiliary learning mechanism is proposed, which not only enables semantic-aware and fine-granular semi-supervised medical image segmentation, but enhances the quality of pseudo labels.
	\item 
	A novel triple-view disparity training strategy is proposed, which uses only three labeled slices of a few volumes to encourage the disparity of sub-networks, significantly improving the robustness.
	\item 
	Extensive experimental results on three challenging semi-supervised segmentation benchmarks, including LA2018, KiTS19, and LiTS, across different modalities (i.e., MR and CT), verify the superiority of our SGTC in comparison with recent state-of-the-art methods.
\end{itemize}

\section{Related Work}
\textbf{Semi-supervised medical image segmentation methods}:
Recently, learning from a constrained pool of labeled data alongside copious amounts of unlabeled data becomes a pragmatic approach in medical image analysis domain~\cite{bai2023bidirectional,cai2021avlsm}. Existing semi-supervised medical image segmentation methods can be classified into two groups: pseudo-label-based methods \cite{wang2022ssa,thompson2022pseudo,uamt,mt} and consistency-based methods~\cite{cps,sassnet,dtc,zhang2024cross,manifold}. The pseudo-label-based methods, like MT~\cite{mt} and UA-MT~\cite{uamt}, estimate the pseudo labels based on a few labeled samples.
Further, BCP \cite{bai2023bidirectional} utilizes a bidirectional copy-paste method to reduce the distribution gap between labeled and unlabeled data. The consistency-based methods, like SASSNet~\cite{sassnet}, CPS~\cite{cps}, and DTC~\cite{dtc}, which employ consistency regularization among different sub-networks. Nevertheless, the above methods still need to fully annotate a volumetric sample, thus limiting their applications in clinical practice. To solve this, methods using sparse annotations, such as PLN~\cite{li2022pln} and Desco~\cite{Desco} utilize the registration methods to propagate few labeled slices to others. Unfortunately, these methods have limitations in semantic understanding, which leads to less accurate boundary segmentation ~\cite{xu2023masqclip}. Besides, due to the absence of information from certain planes, the above methods fails to fully utilize the different planes in the 3D space, thus making it ineffective in modeling the complex distribution of the entire volume.   

\begin{figure*}[!ht]
\footnotesize
\centering
\setlength{\abovecaptionskip}{0cm}
\includegraphics[width=0.98\linewidth]{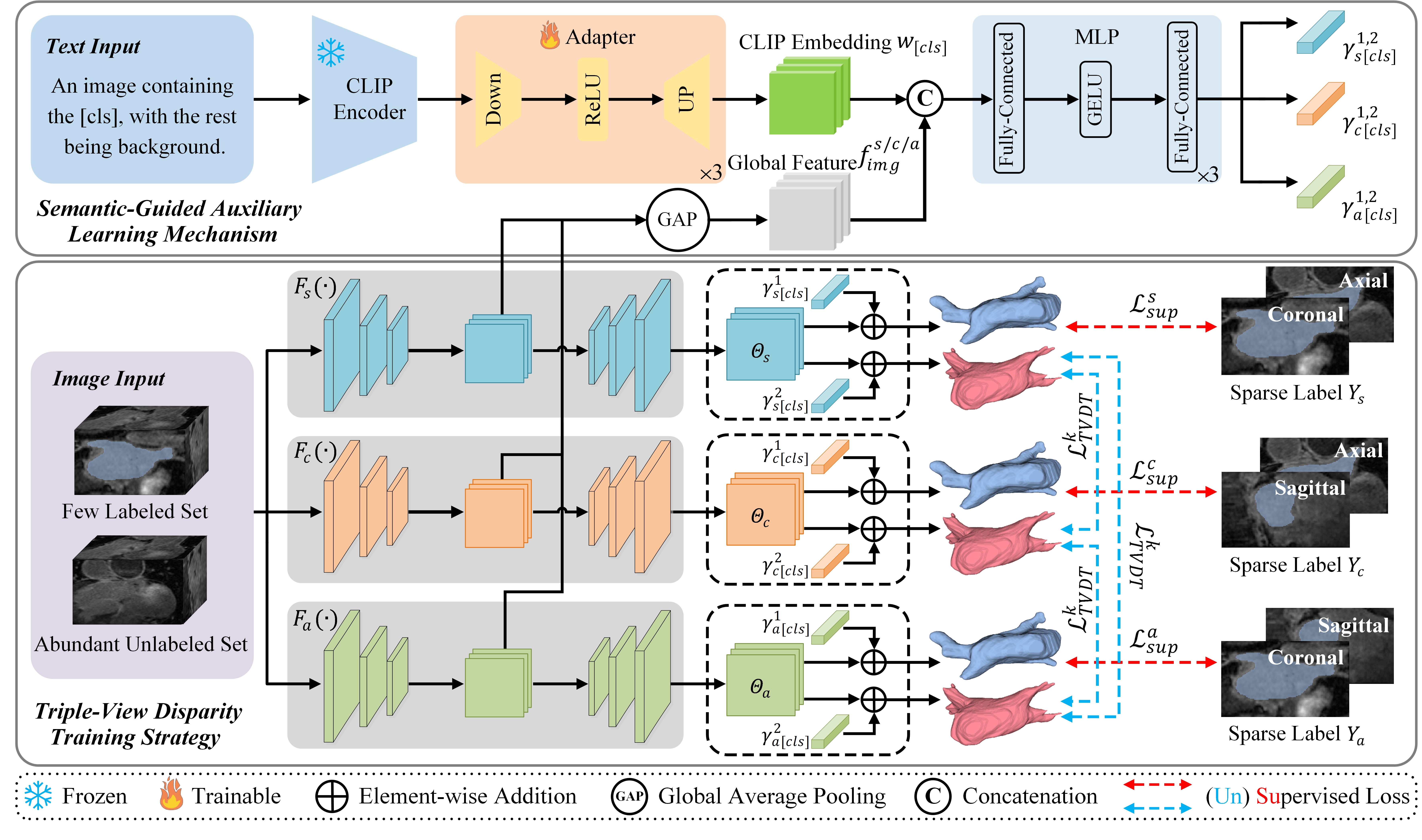}
\caption{Architecture of the proposed SGTC framework. For volumes with sparse orthogonal labels, each volume has three corresponding labels. For model $F_s(\cdot)$, the supervision signals are selected from the Coronal and Axial plane, for model $F_c(\cdot)$ from the Sagittal and Axial plane, and for model $F_a(\cdot)$ from the Coronal and Sagittal planes. For volumes without labels, each segmentation result of $F_s(\cdot)$, $F_c(\cdot)$, and $F_a(\cdot)$ act as cross-supervision signals for other sub-networks.}
\label{fig_3}
\end{figure*}

\textbf{Text-guided methods}: Contrastive Language-Image Pre-training (CLIP~\cite{radford2021learning}) is gaining popularity and has achieved impressive results in various downstream tasks~\cite{guo2023clip,wang2023transferring,yu2022towards,wang2023hierarchical,li2024pp}. Particularly, in the medical image segmentation domain, accurately extracting intricate structures and delineating weak boundaries typically hinges on understanding the semantic nuances within the images. Nowadays, some researchers have started to investigate cross-modal networks in the medical imaging community \cite{chen2022align,cong2022caption,yuan2023ramm}. For instance, Huang \textit{et al.} \cite{huang2021gloria} learns global and local representations of images by comparing subregions of images with words from medical reports. Li \textit{et al.} \cite{li2023lvit} establish a multi-modal dataset containing X-ray and CT images, supplemented with medical text annotations to address quality issues in manually annotated images. Moreover, Liu \textit{et al.} \cite{liu2023clip} adopt the text embedding extracted by CLIP as parameters on the image features. Exploring these foundation models for data-efficient medical image segmentation is still limited, but is highly necessary. To this end, this paper makes one of the first attempts to propose a novel semantic-guided triple co-training framework, which leverages text representations to enhance semi-supervised learning to harness more discriminative semantic information, achieving semantic-aware and fine-granular semi-supervised medical image segmentation.

\section{Methodology}
\subsection{Preliminaries}
In the following, we first define some related preliminaries of this work. Specifically, given a training dataset $D$, we have a labeled set containing $M$ labeled cases, represented as $D_{L}=\{(X_{i}^{l},Y_{i})\}_{i=1}^{M}$, where $X_{i}^{l}$ denotes the input images and $y_{i}$ denotes the corresponding ground truth. Additionally, the unlabeled set $D_{U}$ containing  $N$ unlabeled cases, represented as $D_{U}=\{X_{i}^{u}\}_{i=1}^{N}$, where $N\gg M$. Our method annotates three slices from three orthogonal planes, including $X_{is}^{lp}$ ($s$ means sagittal), $X_{ic}^{lq}$($c$ means coronal), and $X_{ia}^{lr}$($a$ means axial). Overall, there are $p^{th}$ slice in plane $s$, $q^{th}$ slice in plane $c$, and $r^{th}$ slice in plane $a$. Their corresponding annotations are $Y_{is}^{p}$, $Y_{ic}^{q}$, and $Y_{ia}^{r}$, respectively.

\subsection{Model Architecture}
Figure \ref{fig_3} illustrates our proposed Semantic-Guided Triplet Co-Training (SGTC) framework, which consists of two main components: semantic-guided auxiliary learning mechanism and triple-view disparity training strategy. Benefiting from these, our SGTC introduces text representations to enhance semi-supervised learning to exploit more discriminative semantic information. Then, we will elaborate on the technical details of each component step by step.
  
\textbf{Semantic-Guided Auxiliary Learning Mechanism (SGAL)}: While some methods have incorporated textual semantics as auxiliary information to guide segmentation, these methods are still fully supervised, requiring a large amount of labeled data, thus limiting their applicability in clinical settings \cite{li2023lvit,liu2023clip,shin2022cosmos}. To fill this blank, we propose a novel semantic-guided auxiliary learning mechanism, utilizing the text representations from the pre-trained CLIP to enable semantic-aware and fine-granular semi-supervised medical image segmentation, while further enhancing the quality of pseudo labels. 

Specifically, the designed medical prompts are processed through the pre-trained text encoder of CLIP to obtain the text embedding $w$. Since the CLIP is pre-trained on natural images (i.e., the domain gap between natural images and medical images~\cite{ye2022alleviating}), it may prevent the CLIP text encoder from fully capturing the clinical semantics contained in medical prompts. Therefore, we freeze the pre-trained CLIP and fine-tune the Adapter module followed by the frozen CLIP encoder. The Adapter module consists of a dimension reduction projection layer followed by an activation function layer, and an up projection layer. The calculation process of $w$ can be written as:
\begin{equation}
    w=\text{Adapter}(\text{CLIP}_{Enc}(\text{Text Prompt})).
\end{equation}

In this paper, the text prompts are defined as ``\textit{An image containing the [CLS], with the rest being background}'', where [CLS] is a concrete class name. It is well known that the template of medical prompts is crucial, and therefore the effectiveness of different prompt templates will be verified in the following studies. After obtaining the text embedding $w$, we concatenate $w$ with the global image feature $f_{img}$ extracted through the encoder path of the segmentation network to better align the image-text modality. Subsequently, this concatenated representation is directly fed into a multi-layer perception (MLP) to obtain the cross-modal parameters $\gamma_{s/c/a}^{1,2}$, which can be formulated as follows:
\begin{equation}
    \gamma_{s/c/a}^{1,2}=\text{MLP}(w~\copyright~ f_{img}^{s/c/a}),
\end{equation}
where $\copyright$ represents concatenation. Then, we perform element-wise addition of the cross-modal parameters $\gamma$ and the features extracted before the final classification layer $\Theta_{s/c/a}$ of each sub-networks, and pass it through a convolution layer to obtain the predictions $P_{s/c/a}$ of each branch.
\begin{equation}
P_{s/c/a}=\text{Conv3D}(\Theta_{s/c/a} \oplus \gamma_{s/c/a}^{1,2}),
\end{equation}
where $\oplus$ represents element-wise addition, and Conv3D denotes the 3D convolution layer.

\textbf{Triple-View Disparity Training Strategy (TVDT)}: The previous sparse annotation approach, which annotated slices on only one or two planes, failed to fully preserve spatial information, leading to a suboptimal segmentation results. To address this issue, we proposed a novel triple-view disparity training strategy, which significantly improves the robustness by maintaining the disparity of sub-networks during training as well as allowing the sub-networks to learn complementary knowledge from each other. More importantly, it just needs the clinician to annotate three orthogonal slices of a few volumetric samples. 

Specifically, for a volume $X^l$, we first employ three distinct sparse labels $Y_{s}$, $Y_{c}$, and $Y_{a}$ to supervise the three sub-networks respectively. This ensures better consistency training while maintaining the disagreement among different sub-networks. For each sub-network, we select two orthogonal annotated slices from the three orthogonal planes as supervision signals. For the sparse label $Y_s$, we choose  $Y_{c}^{q}$ slice and $Y_{a}^{r}$ slice, which can be formulated as follows:
\begin{equation}
Y_{s}=Y \otimes W_{s},
\end{equation}
\begin{equation}W_s^i=\begin{cases}1,&\quad\text{if voxel $i$ is on the selected slices,}\\0,&\quad\text{otherwise},\end{cases}\end{equation}
where $\otimes$ denotes element-wise multiplication. $W_s$ means the weight matrix. Similarly, for $Y_c$, we choose $Y_{s}^{p}$ and $Y_{a}^{r}$ slice, and for $Y_a$, $Y_{s}^{p}$ and $Y_{c}^{q}$ slice are allocate.

Our SGTC comprises three 3D segmentation networks (i.e., Vnet~\cite{vnet}), denoted as $F_s(\cdot)$, $F_c(\cdot)$, and $F_a(\cdot)$. As mentioned above, every volume $X_i^l$ ($i\le M$) merely has three orthogonal labels $Y_s$ ,$Y_c$ and $Y_a$. Sub-network $F_s(\cdot)$ is trained with $Y_s$, $F_c(\cdot)$ is trained with $Y_c$, and $F_a(\cdot)$ is trained with $Y_a$, respectively. Under this setting, all three sub-networks can learn knowledge from different planes of the others while maintaining key spatial information of 3D medical volumes.

For abundant unlabeled data, the triplet sub-networks guide each other in turn, where the predictions generated by model $F_s(\cdot)$ serve as pseudo labels for $F_c(\cdot)$ and $F_a(\cdot)$. Here, following \cite{uamt}, we compute uncertainty using Monte Carlo dropout, selecting voxels with uncertainties lower than a threshold as better pseudo labels for triplet co-training. Similarly, the predicted results of $F_c(\cdot)$ and $F_a(\cdot)$ serve as pseudo labels for the other two sub-networks. The loss is computed using weighted cross-entropy loss as:
\begin{equation}
\mathcal{L}_{TVDT}^{k}=-\frac{1}{\sum_{i=1}^{H \times W \times D} m_i} \sum_{i=1}^{H \times W \times D} m_i \hat{y}_{i}^{k} \log p_i, k = 1,2,
\end{equation}
where $m_i$ denotes whether the $i^{th}$ voxel is selected. $p_{i}$ is the prediction result of the current model, and $\hat{y}_{i}^{1}$ and $\hat{y}_{i}^{2}$ are the pseudo labels generated by the other two sub-networks.

In addition, for limited labeled data, the supervised loss contains weighted cross-entropy loss and weighted dice loss, which can be defined as follows:
\begin{equation}
 \mathcal{L}_{WCE}=-\frac{1}{\sum_{i=1}^{H\times W\times D}w_i}\sum_{i=1}^{H\times W\times D}w_iy_i\log p_i,
\end{equation}

\begin{equation}
 \mathcal{L}_{Dice}=1-\frac{2\times\sum_{i=1}^{H\times W\times D}w_ip_iy_i}{\sum_{i=1}^{H\times W\times D}w_i(p_i^2+y_i^2)},
\end{equation}
where $w_i$ is the $i^{th}$ voxel value of the weight matrix. $p_i$ and $y_i$ respectively represent the predicted results and the ground truth labels of the network on voxel $i$. Therefore, the supervised learning loss is represented as follows:
\begin{equation}
 \mathcal{L}_{Sup}=\frac{1}{2}\mathcal{L}_{WCE}+\frac{1}{2}\mathcal{L}_{Dice}.
\end{equation}

Overall, the total loss during training is defined as:
\begin{equation}
\label{equ_10}
\mathcal{L}_{SGTC}=(1-\alpha)\mathcal{L}_{Sup}+\alpha\mathcal{L}_{TVDT}^{k},
\end{equation}
where the first term $\mathcal{L}_{Sup}$ is tailored for labeled data, and the second term $\mathcal{L}_{TVDT}^{k}$ is employed for unlabeled data. $\alpha$ denotes the proposed dynamic parameter, which enables the supervised loss from the three annotated slices to dominate the training at the first few epochs, and then gradually increase the weight of the unsupervised loss to stabilize the entire semi-supervised learning. 

\begin{table*}[!t]
    \footnotesize
    \centering
    \renewcommand{\arraystretch}{0.85} 
    \setlength{\abovecaptionskip}{0.1cm}
    \resizebox{\textwidth}{!}{
        \begin{tabular}{p{7cm}ccccccc}
            \toprule
            \multirow{2}{*}{Method} & \multirow{2}{*}{Labeled Slices} & \multicolumn{2}{c}{Scans Used}  & \multicolumn{4}{c}{Metrics}                               \\ \cmidrule(l){3-8} 
             & & Labeled       & Unlabeled      & Dice $\uparrow$          & Jaccard $\uparrow$           & HD $\downarrow$ & ASD $\downarrow$ \\ \midrule
     
            MT \cite{mt} (NIPS'17)    &CA            & 8             & 72             & 0.661 $\pm$ 0.124 & 0.505 $\pm$ 0.136               & 38.681 $\pm$ 9.600           & 13.597 $\pm$ 3.730  \\
            UA-MT \cite{uamt} (MICCAI'19)   &CA                 & 8             & 72             & 0.650 $\pm$ 0.096        & 0.489 $\pm$ 0.108       & 40.442 $\pm$ 8.739           & 14.841 $\pm$ 4.041      \\
            SASSNet \cite{sassnet} (MICCAI'20)   &CA         & 8             & 72             & 0.617 $\pm$ 0.119      & 0.456 $\pm$ 0.121              & 41.913 $\pm$ 9.348           & 15.521 $\pm$ 4.717     \\
           CPS \cite{cps} (CVPR'21)    &CA      & 8             & 72             &0.661 $\pm$ 0.082      & 0.499 $\pm$ 0.092              & 38.718 $\pm$ 6.958           & 13.992 $\pm$ 3.158        \\
           DTC \cite{dtc} (AAAI'21)   &CA       & 8             & 72             & 0.686 $\pm$ 0.120     & 0.533 $\pm$ 0.133              & 35.624 $\pm$ 8.651           & 11.556 $\pm$ 4.035        \\
          BCP \cite{bai2023bidirectional} (CVPR'23)     &CA       & 8             & 72             & 0.784 $\pm$ 0.093      & 0.653 $\pm$ 0.115            & 22.432 $\pm$ 8.282           & 5.753 $\pm$ 2.667         \\
           Desco \cite{Desco} (CVPR'23)     &CA       & 8             & 72             & 0.711 $\pm$ 0.093      & 0.559 $\pm$ 0.111             & 35.671 $\pm$ 7.134           & 12.776 $\pm$ 3.532         \\ 

           SGTC (Dual)       &CA       & 8             & 72             & 0.739 $\pm$ 0.078       & 0.592 $\pm$ 0.097              & 35.464 $\pm$ 9.110          & 11.503 $\pm$ 4.015     \\
           SGTC (Ours)      &SCA     & 8             & 72             & \textbf{0.847 $\pm$ 0.044}      & \textbf{0.738 $\pm$ 0.066}              &\textbf{ 17.442 $\pm$ 11.719}           & \textbf{ 4.256 $\pm$ 3.836}      
                 \\ \bottomrule
        \end{tabular}
        }
     \caption{Quantitative comparisons with the seven state-of-the-art methods on the LA2018 dataset under 10\%  labeled cases. In this paper, \textbf{bold} values denote the best-performing method. S, C, and A indicate sagittal, coronal, and axial annotated slices.}
        \label{table1}
    \end{table*}

\begin{figure*}[!t]
\footnotesize
\centering
\setlength{\abovecaptionskip}{0cm}
\includegraphics[width=1.0\linewidth]{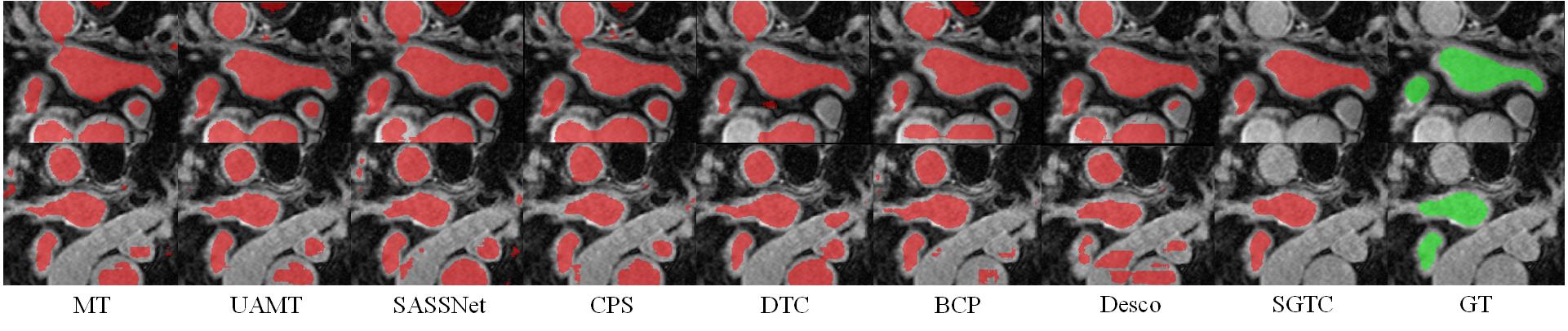}
\caption{Qualitative comparisons on the LA2018 dataset with 10\% labeled cases. From left to right: segmentation results of MT~\cite{mt}, UAMT~\cite{uamt}, SASSNet~\cite{sassnet}, CPS~\cite{cps},
DTC~\cite{dtc}, BCP~\cite{bai2023bidirectional}, Desco~\cite{Desco}, our SGTC and ground truth (GT), respectively.}
\label{fig_4}
\end{figure*}

\section{Experimental Results}
\subsection{Datasets \& Metrics}

\textbf{LA2018 Dataset} \cite{xiong2021global} contains 100 gadolinium enhanced MR imaging scans with labels. All scans have the same isotropic resolution of $0.625\times0.625\times0.625mm^3$. Following existing works \cite{uamt,dtc}, we utilize 80 training samples and 20 testing samples for fair comparison with other methods.

\textbf{KiTS19 Dataset} \cite{heller2019kits19} 
 is provided by the Medical Centre of Minnesota University and consists of 300 abdominal CT scans. The slice thickness ranges from 1mm to 5mm. The dataset is divided into 190 training samples and 20 testing samples.

 \textbf{LiTS Dataset} \cite{bilic2023liver} 
 is a CT dataset focused on liver and liver tumour segmentation. The dataset collects 201 abdominal scans. Among these, 131 scans with segmentation masks are publicly available. We use the same split of the dataset as done in \cite{Desco}, where 100 scans are used for training, and the remaining 31 scans for test.

To fairly compare our method with others, we adopt four commonly used metrics: Dice similarity coefficient (Dice), Jaccard similarity coefficient (Jaccard), 95\% Hausdorff Distance (95HD), and Average Surface Distance (ASD).

\begin{table*}[!t]
    \footnotesize
    \centering
    \renewcommand{\arraystretch}{0.85} 
    \setlength{\abovecaptionskip}{0.1cm}
    \resizebox{\textwidth}{!}{
        \begin{tabular}{p{7cm}ccccccc}
            \toprule
            \multirow{2}{*}{Method} & \multirow{2}{*}{Labeled Slices} & \multicolumn{2}{c}{Scans Used} &  \multicolumn{4}{c}{Metrics}                           \\ \cmidrule(l){3-8} 
             & & Labeled       & Unlabeled      & Dice $\uparrow$          & Jaccard $\uparrow$           & HD $\downarrow$ & ASD $\downarrow$  \\ \midrule
     
            MT \cite{mt} (NIPS'17)     &CA          & 19             & 171             & 0.780 $\pm$ 0.084 & 0.647 $\pm$ 0.113               & 43.797 $\pm$ 13.586           & 16.152 $\pm$ 6.264  \\
            UA-MT \cite{uamt} (MICCAI'19)   & CA             & 19             & 171             & 0.856 $\pm$ 0.096        & 0.758 $\pm$ 0.137       & 32.081 $\pm$ 15.390           & 9.120 $\pm$ 4.739    \\
            SASSNet \cite{sassnet} (MICCAI'20)  &CA            & 19             & 171             & 0.914 $\pm$ 0.054      & 0.845 $\pm$ 0.087              & 22.130 $\pm$ 15.827           & 5.856 $\pm$ 3.551    \\
           CPS \cite{cps} (CVPR'21)    &CA       & 19             & 171             &0.813 $\pm$ 0.106      & 0.697 $\pm$ 0.146              & 42.438 $\pm$ 9.210          & 13.754 $\pm$ 5.482      \\
           DTC \cite{dtc} (AAAI'21)   &CA          & 19            & 171             & 0.918 $\pm$ 0.058     & 0.853 $\pm$ 0.094              & 12.049 $\pm$ 16.383           & 3.174 $\pm$ 3.084       \\
          BCP \cite{bai2023bidirectional} (CVPR'23)     &CA         & 19             & 171             & 0.921 $\pm$ 0.046      & 0.864 $\pm$ 0.083              & 6.726 $\pm$ 11.021           &  2.888 $\pm$ 3.807         \\
           Desco \cite{Desco} (CVPR'23)     &CA         & 19             & 171             & 0.880 $\pm$ 0.099      & 0.798 $\pm$ 0.145              & 21.567 $\pm$ 18.906           & 6.257 $\pm$ 4.967         \\

          SGTC (Dual)   &CA           & 19            & 171            & 0.927 $\pm$ 0.059      & 0.870 $\pm$ 0.092              & 7.821 $\pm$ 9.638           & 2.639 $\pm$ 1.996   \\ 
           SGTC (Ours)   &SCA       & 19            & 171            & \textbf{ 0.933 $\pm$ 0.041 }     & \textbf{ 0.877 $\pm$ 0.067 }             & \textbf{ 5.145 $\pm$ 6.983 }          & \textbf{ 2.038 $\pm$ 1.868 }    
                 
                 \\ \bottomrule
        \end{tabular}
        }
        \caption{Quantitative comparisons with the seven state-of-the-art methods on the KITS19 dataset under 10\% labeled cases.} 
        \label{table2}
    \end{table*}

\begin{figure*}[!t]
\footnotesize
\centering
\setlength{\abovecaptionskip}{0cm}
\includegraphics[width=1.0\linewidth]{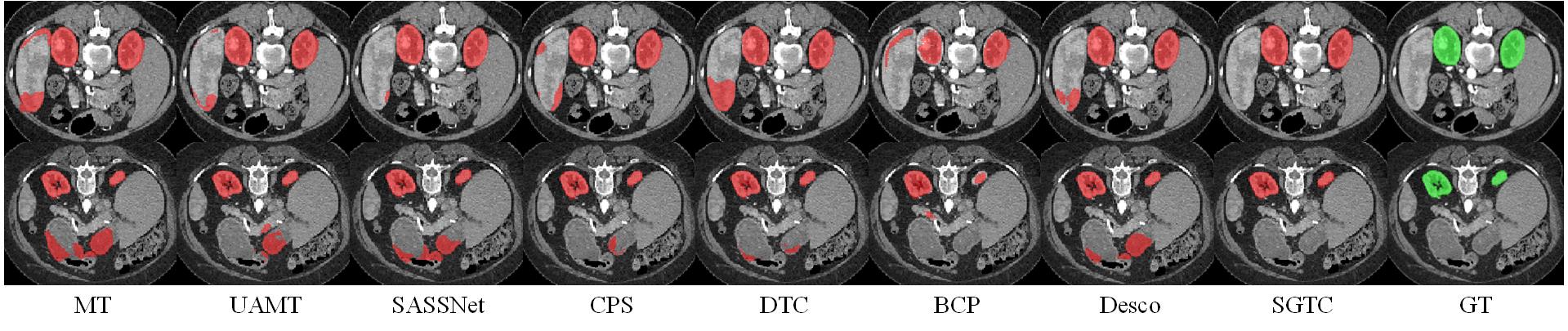}
\caption{Qualitative comparisons on the KITS19 dataset with 10\% labeled cases.}
\label{fig_5}
\end{figure*}

\begin{table*}[!ht]
    \footnotesize
    \centering
    \setlength{\abovecaptionskip}{0.1cm}
    \renewcommand{\arraystretch}{0.85} 
    \resizebox{\textwidth}{!}{
        \begin{tabular}{p{7cm}ccccccc}
            \toprule
            \multirow{2}{*}{Method} & \multirow{2}{*}{Labeled Slices} & \multicolumn{2}{c}{Scans Used} & \multicolumn{4}{c}{Metrics}  \\ \cmidrule(l){3-8} 
             &  & Labeled       & Unlabeled      & Dice $\uparrow$          & Jaccard $\uparrow$           & HD $\downarrow$ & ASD $\downarrow$ \\ \midrule
     
            MT \cite{mt} (NIPS'17)    &CA           & 10             & 90             & 0.829 $\pm$ 0.114 & 0.721 $\pm$ 0.137               & 46.629 $\pm$ 20.668           & 13.122 $\pm$ 7.901  \\
            UA-MT \cite{uamt} (MICCAI'19)    &CA             & 10             & 90             & 0.781 $\pm$ 0.212       & 0.677 $\pm$ 0.221       & 20.097 $\pm$ 15.522           & 5.277 $\pm$ 4.343      \\
            SASSNet \cite{sassnet} (MICCAI'20)    &CA         & 10             & 90             & 0.830 $\pm$ 0.119      & 0.724 $\pm$ 0.143              & 46.115 $\pm$ 12.276           & 13.822 $\pm$ 5.276    \\
           CPS \cite{cps} (CVPR'21)  &CA          & 10             & 90             &0.827 $\pm$ 0.089      & 0.713 $\pm$ 0.122              & 47.098 $\pm$ 11.201          & 12.521 $\pm$ 4.162     \\
           DTC \cite{dtc} (AAAI'21)    &CA      & 10            & 90             & 0.896 $\pm$ 0.071     & 0.817 $\pm$ 0.107              & 18.926 $\pm$ 16.982           & 5.519 $\pm$ 4.432         \\
           BCP \cite{bai2023bidirectional} (CVPR'23)    &CA         & 10             & 90            &  0.922 $\pm$ 0.060     & 0.860 $\pm$ 0.094              & 11.224 $\pm$ 15.202          &3.294 $\pm$ 4.121        \\
           Desco \cite{Desco} (CVPR'23)    &CA         & 10             & 90            & 0.885 $\pm$ 0.055      & 0.798 $\pm$ 0.083              & 16.095 $\pm$ 11.683          &4.323 $\pm$ 3.501        \\

             SGTC (Dual)     &CA         & 10             & 80             & 0.919 $\pm$ 0.049      &0.854 $\pm$ 0.079             & 10.960 $\pm$ 13.698         &3.166 $\pm$ 3.926     
     \\
           SGTC (Ours)    &SCA    & 10            & 90           & \textbf{ 0.927 $\pm$ 0.044 }     & \textbf{ 0.867 $\pm$ 0.071 }             & \textbf{ 9.302 $\pm$ 11.694 }          & \textbf{ 2.710 $\pm$ 3.104 }   
                 \\
             \bottomrule
        \end{tabular}
        }
        \caption{Quantitative comparisons with the seven state-of-the-art methods on the LITS dataset under 10\%  labeled cases.}
        \label{table3}
    \end{table*}

\begin{figure*}[!t]
\footnotesize
\centering
\setlength{\abovecaptionskip}{0cm}
\includegraphics[width=1.0\linewidth]{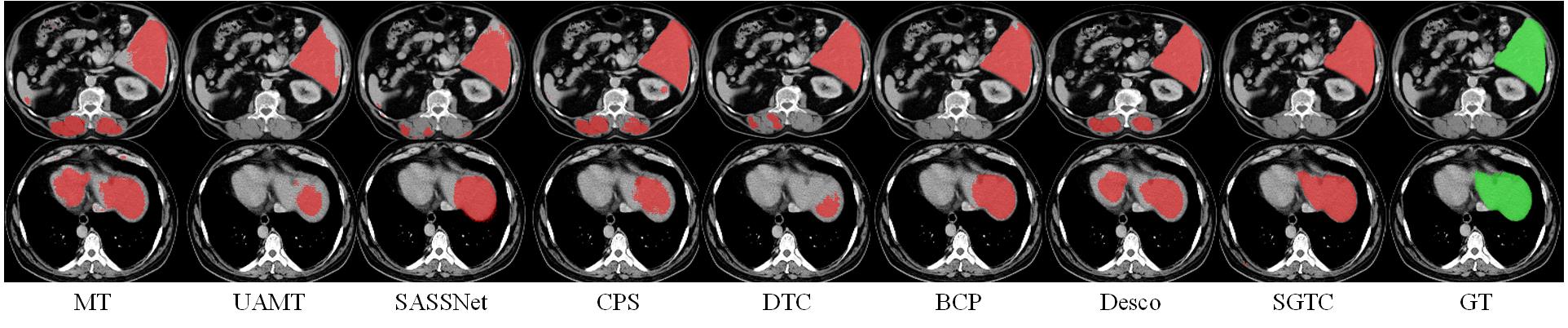}
\caption{Qualitative comparisons on the LITS dataset with 10\% labeled cases.}
\label{fig_6}
\end{figure*}

\subsection{Implementation Details}
To make sparse annotations more effective, the three selected slices should contain the foreground area of the segmentation target, and we choose slices as close to the center position as possible in all three planes. The entire training is conducted in the PyTorch framework. We set the batch size to 4, with each batch containing two volumes with labels and two volumes without labels. We train for 6000 iterations using the Stochastic Gradient Descent (SGD) optimizer. The initial learning rate is set to 0.01 and gradually decays to 0.0001. The value of parameter $\alpha$ is initialized to 0.1 and is increased every 150 iterations.

\subsection{Comparison Experiments}
In this paper, we compare our SGTC with 7 recent SOTAs semi-supervise medical image segmentation methods, including MT~\cite{mt}, UA-MT~\cite{uamt}, SASSNet~\cite{sassnet}, CPS~\cite{cps}, DTC~\cite{dtc}, BCP~\cite{bai2023bidirectional} and Desco~\cite{Desco}   on three challenging benchmarks, i.e., LA2018~\cite{xiong2021global}, KiTS19~\cite{heller2019kits19}, and LiTS~\cite{heller2019kits19} datasets.
In other methods and the SGTC (Dual) version, we annotate two slices per volume according to the previous settings \cite{Desco}. In the SGTC version, we annotate three slices per volume. The S, C, A indicates whether the annotated slices is sagittal, coronal or axial plane.

\textbf{Comparison Results on LA2018 dataset}: Table~\ref{table1} shows the comparison between our method and these state-of-the-art models on LA2018 dataset under 10\% (8 samples), which shows our method achieves the best performance (i.e., surpassing the second best by
6.3\% on Dice). Due to the semantic guidance, the network can comprehensively perceive subtle changes in boundary regions and voxel-level semantic information, which makes the network perform better in dealing with complex anatomy structures. Additionally, methods using two orthogonal annotations generally perform poorly, whereas our SGTC with three orthogonal annotations achieves better shape-related performance by modeling the entire volume's supervision signal distribution more comprehensively. Visual comparisons are shown in Figure \ref{fig_4}, where our SGTC accurately delineates the intricate structures and weak boundaries. More visualizations can be found in the supplementary materials.

\textbf{Comparison Results on KITS19 dataset}: Table \ref{table2} demonstrates the comparison between our method and recent SOTAs on KITS19 dataset under 10\% (19 samples) labeled data settings. We can find that our SGTC outperforms existing methods in all metrics. For example, compared with BCP~\cite{bai2023bidirectional} and DTC~\cite{dtc}, SGTC shows 1.2\% and 1.5\% improvements on Dice. Besides, the qualitative comparison results are shown in Figure \ref{fig_5}, which shows our method effectively handles complex boundaries and fine structures.

\textbf{Comparison Results on LITS dataset}: Table \ref{table3} presents the comparison between our method and others on the LITS dataset under 10\% (10 samples) labeled data settings. It can be observed that our method's advantage becomes more pronounced with fewer labeled data. The main reason behind this is that our method achieves semantic-aware and fine-granular segmentation.
Moreover, visual comparison results in Figure \ref{fig_6} that show our SGTC effectively handles complex boundaries and fine structures.

\subsection{Ablation Study and Analysis}

\textbf{Ablation study of each component}:
As shown in Table \ref{table4}, we conduct an ablation experiment on the KiTS19 dataset with 9 volumes to  better evaluate the components of our method. By incorporating semantic cues, our method is able to focus more effectively on boundary information, leading to improved performance. 

\begin{table}[!t]
    \footnotesize
    \centering
    \setlength{\abovecaptionskip}{0.1cm}
    \renewcommand{\arraystretch}{0.7} 
    
    \begin{centering}
		\setlength{\tabcolsep}{2mm}{
        \begin{tabular}{cccccc}
            \toprule
             SGAL & TVDT & Dice $\uparrow$ &Jaccard $\uparrow$ &HD $\downarrow$ &ASD $\downarrow$  \\ \midrule
             $\times$ &$\times$ &0.780  & 0.655 & 41.630 &14.110     \\
            \checkmark &$\times$ & 0.892 &0.814 & 10.945 & 3.318\\
            $\times$ &\checkmark  & 0.896 & 0.817 & 11.985 & 3.106 \\
            \checkmark     &\checkmark  &\textbf{0.915} &   \textbf{0.837 }&\textbf{9.968}   & \textbf{2.668} \\
                 \bottomrule
        \end{tabular}}
        \caption{Ablation study of on KiTS19 dataset.}
		\label{table4}
	 \end{centering}
    \end{table}

\textbf{Ablation study of triple-view disparity training strategy}: As shown in Table \ref{table5}, where all these methods use the same annotated strategy (i.e., CAC or SCA). The results indicate that our method outperforms others under the same settings, and SCA provides better performance improvement, demonstrating that the performance gains are due to the more effective triple-view disparity training strategy.

\textbf{Ablation study of text prompts in CLIP}: Table \ref{table6} illustrates the results of using different text prompts. It can be observed that, compared to solely using image features, combining visual and textual modalities brings large performance gains. Our tailored text prompt, which effectively integrates semantics into the network by describing more details, resulting in more promising outcomes.

\begin{table}[!t]
    \footnotesize
    \centering
    \renewcommand{\arraystretch}{0.7} 
    \setlength{\abovecaptionskip}{0.1cm}
      
    \begin{centering}
		\setlength{\tabcolsep}{0.5mm}{
        \begin{tabular}{ccccc}
            \toprule
            \multirow{2}{*}{Method} & \multirow{2}{*}{Labeled Slices} & \multicolumn{2}{c}{Scan Used} & \multirow{2}{*}{Dice} \\ \cmidrule(lr){3-4}
           &  & Labeled       & Unlabeled      &      \\ \midrule
     
            MT &CAC / SCA                & 9             & 181             & 0.782 / 0.809  \\
            UA-MT     &CAC / SCA            & 9             & 181             & 0.851 / 0.871            \\
            SASSNet        &CAC / SCA       & 9             & 181             & 0.885 / 0.897         \\
           CPS            &CAC / SCA    & 9             & 181             &0.826 / 0.841          \\
           DTC          &CAC / SCA     & 9             & 181            & 0.887 / 0.896          \\
           BCP        &CAC / SCA        & 9             & 181             & 0.909 / 0.913        \\
           Desco        &CAC / SCA        & 9             & 181             & 0.828 / 0.855          \\

           SGTC (Ours)             &CAC / SCA & 9             & 181             & \textbf{ 0.911 / 0.915 }     
                 \\ \bottomrule
        \end{tabular}}
        \caption{Ablation study of the proposed triple-view disparity training strategy on KITS19 dataset.}
		\label{table5}
	 \end{centering}
    \end{table}

\begin{table}[!t]    
\footnotesize
    \centering 
        \setlength{\abovecaptionskip}{0.1cm}
    \renewcommand{\arraystretch}{0.1} 
    
    \begin{centering}
		 \setlength{\tabcolsep}{1.0mm}{
        \begin{tabular}{cccc}
            \toprule
            \multirow{2}{*}{Text prompts} & \multicolumn{2}{c}{Scan Used} & \multirow{2}{*}{Dice} \\ \cmidrule(lr){2-3}
             & Labeled       & Unlabeled      &      \\ \midrule
     
            \makecell{None.}                   & 6             & 184             & 0.873  \\ \cdashline{1-4}
            \makecell{A photo of a [cls].}                   & 6             & 184             & 0.882        \\ \cdashline{1-4} 
            \makecell{There is a [cls] in this \\computerized 
            tomography/\\magnetic resonance imaging.}            & 6             & 184             &  0.883         \\ \cdashline{1-4}
           \makecell{An image containing the [cls],\\
        with the rest being background.}              & 6             & 184             & \textbf{0.888 }  

                 \\ \bottomrule
        \end{tabular}}
        \caption{Ablation study of different text prompts.}

		\label{table6}
	 \end{centering}
    \end{table}
\begin{figure}[!t]
\footnotesize
\centering
\setlength{\abovecaptionskip}{0cm}
  \includegraphics[width=\linewidth]{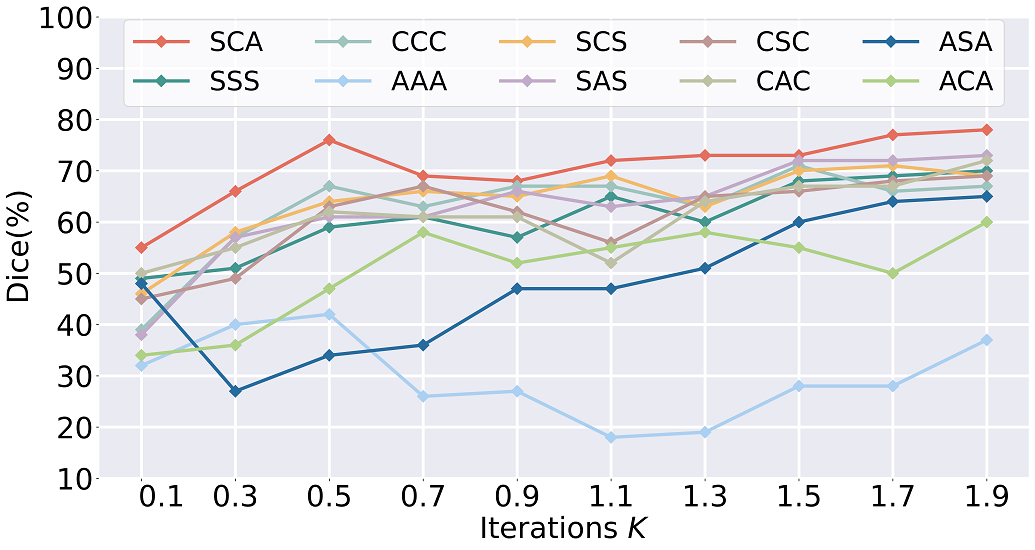}
  \caption{Performance comparisons of different annotations.}
  \label{fig_8}
\end{figure}
\textbf{Ablation study of the different annotation method}: Figure \ref{fig_8} illustrates segmentation performance by using our proposed three orthogonal annotations, the two orthogonal annotations  (i.e., the third slice is parallel to one of the first two) and the parallel annotations (i.e., three parallel slices in the same plane). It is shown that our annotation scheme achieves superior results at different iterations. Additionally, Figure \ref{fig_9} shows t-SNE visualization of the extracted features using different annotation methods. Specifically, we trained five networks with  just one single annotated slice per volume: three orthogonal slices for training $s$, $c$, and $a_1$, and three parallel slices for training $a_1$, $a_2$, and $a_3$. The features extracted using three orthogonal annotations are more concentrated, demonstrating the effectiveness of our triple-view disparity training strategy.

\begin{figure}[!t]
\footnotesize
\centering
\setlength{\abovecaptionskip}{0cm}
  \includegraphics[width=0.95\linewidth]{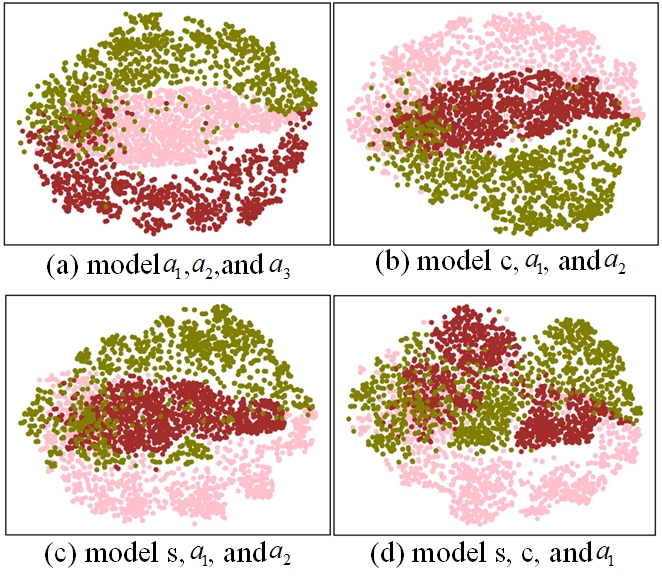}
  \caption{The t-SNE visualization of feature representations extracted from models trained on (a) three parallel slices, (b), (c) two orthogonal slices. (d) three orthogonal slices.}
  \label{fig_9}
\end{figure}

\begin{table}[!t]
    \footnotesize
    \centering
    \setlength{\abovecaptionskip}{0.1cm}
    \renewcommand{\arraystretch}{0.7} 
     
    \begin{centering}
		\setlength{\tabcolsep}{2mm}{
        \begin{tabular}{ccccc}
            \toprule
            \multirow{2}{*}{Hyper-Parameter} &  \multicolumn{2}{c}{Scan Used} & \multirow{2}{*}{Dice} &
            \multirow{2}{*}{Jaccard} \\ \cmidrule(lr){2-3}
            & Labeled       & Unlabeled      &      \\ \midrule
     
            $\alpha$ = 0.1     & 38             & 152             & 0.942  &0.891 \\
            $\alpha$ = 0.15    &38             & 152               & 0.944      &0.895      \\
            $\alpha$ = 0.2       & 38             & 152             & 0.945  &0.897       \\
           $\alpha$ = 0.25         & 38             & 152             &0.929 &0.870          \\
          $\alpha$ = 0.3          &38            & 152            & 0.936 &0.882        \\
           The proposed $\alpha$        &38 &152    &\textbf{0.947}             & \textbf{0.901}     
                 \\ \bottomrule
        \end{tabular}}
        \caption{Anaysis of hyper-parameter $\alpha$.}
		\label{table7}
	 \end{centering}
    \end{table}
\textbf{Parameter Analysis}: Table \ref{table7} shows the analysis of the hyper-parameter $\alpha$ in Eq.~\ref{equ_10}. Specifically, we conducted ablation experiments on the KITS19 dataset by setting $\alpha$ to 0.1, 0.15, 0.2, 0.25, 0.3, and the proposed dynamic coefficient. It is observed that the proposed dynamic coefficient yields the best performance, as it makes the training more stable.

\section{Conclusion}
This paper presents a novel Semantic-Guided Triplet Co-training framework (SGTC), for accurate and clinically realistic semi-supervised medical image segmentation, which consists of two major contributions. The proposed semantic-guided auxiliary learning mechanism that generates high-quality pseudo labels for semantic-aware and fine-granular segmentation. The proposed triple-view disparity training strategy that requires annotating  only three orthogonal slices, enhancing sub-network diversity and robustness. Extensive experiments and ablations conducted on three challenging benchmarks demonstrate the effectiveness of our proposed SGTC, showcasing its superiority over most state-of-the-art methods. \textbf{Limitations.} Our SGTC exhibits performance degradation when the selected slices contain limited foreground information. In future work, we plan to optimize our approach to address this limitation.

\section{Acknowledgments}
This work was supported in part by the National Science Foundation of China under Grant62471448, 62102338; in part by  Shandong Provincial Natural Science Foundation under Grant ZR2024YQ004; in part by TaiShan Scholars Youth Expert Program of Shandong Province under Grant No.tsqn202312109.
\renewcommand{\bibfont}{\small}
\bibliography{aaai25_base}
\end{document}